\documentclass[lettersize,journal]{IEEEtran}
\usepackage{amsmath,amsfonts}
\usepackage{algorithmic}
\usepackage{array}
\usepackage[caption=false,font=normalsize,labelfont=sf,textfont=sf]{subfig}
\usepackage{textcomp}
\usepackage{stfloats}
\usepackage{url}
\usepackage{verbatim}
\usepackage{graphicx}
\def\BibTeX{{\rm B\kern-.05em{\sc i\kern-.025em b}\kern-.08em
    T\kern-.1667em\lower.7ex\hbox{E}\kern-.125emX}}
\usepackage{balance}
\usepackage{booktabs}
\usepackage{multirow}
\usepackage{amsmath}
\usepackage{hyperref}
\usepackage{xcolor}

\begin{document}

\title{\LARGE \bf Terrain-Aware Stride-Level Trajectory Forecasting for a Powered Hip Exoskeleton via Vision and Kinematics Fusion
}

\author{{Ruoqi Zhao}\textsuperscript{1}, {Xingbang Yang}\textsuperscript{1,*} and {Yubo Fan}\textsuperscript{*}
\thanks{\textsuperscript{1} These authors contribute equally to this letter.}
\thanks{\textsuperscript{*} Corresponding authors.}
\thanks{Manuscript received xx XXXX XXXX; accepted xx XXXX XXXX. Date of
publication xx XXXX XXXX; date of current version xx XXXX XXXX. This
letter was recommended for publication by Associate Editor XXX and
Editor XXX upon evaluation of the reviewers’ comments. This work was supported by the Beijing Natural Science Foundation under Grant L222139, the Fundamental Research Funds for the Central Universities under Grants YWF-23-Q-1031 and JKF-YG-22-B010. }
\thanks{This work involved human subjects or animals in its research. Approval of all ethical and experimental procedures and protocols was granted by Beihang University Ethics Committee under No. BM20220181.}
\thanks{The authors are with School of Biological Science and Medical Engineering, Key Laboratory of Biomechanics and Mechanobiology, Beihang University, Beijing, 100191, China. (Email: zhaoruoqi@buaa.edu.cn; yangxingbang@buaa.edu.cn; yubofan@buaa.edu.cn)}%
}

\maketitle

\begin{abstract}

Powered hip exoskeletons have shown the ability for locomotion assistance during treadmill walking. However, providing suitable assistance in real-world walking scenarios which involve changing terrain remains challenging. Recent research suggests that forecasting the lower limb joint's angles could provide target trajectories for exoskeletons and prostheses, and the performance could be improved with visual information. In this letter, We share a real-world dataset of 10 healthy subjects walking through five common types of terrain with stride-level label. We design a network called Sandwich Fusion Transformer for Image and Kinematics (SFTIK), which predicts the thigh angle of the ensuing stride given the terrain images at the beginning of the preceding and the ensuing stride and the IMU time series during the preceding stride. We introduce width-level patchify, tailored for egocentric terrain images, to reduce the computational demands. We demonstrate the proposed sandwich input and fusion mechanism could significantly improve the forecasting performance. Overall, the SFTIK outperforms baseline methods, achieving a computational efficiency of 3.31 G Flops, and root mean square error (RMSE) of 3.445 \textpm \ 0.804\textdegree \ and Pearson's correlation coefficient (PCC) of 0.971 \textpm\ 0.025. The results demonstrate that SFTIK could forecast the thigh's angle accurately with low computational cost, which could serve as a terrain adaptive trajectory planning method for hip exoskeletons. Codes and data are available at \href{https://github.com/RuoqiZhao116/SFTIK}{https://github.com/RuoqiZhao116/SFTIK}.

\end{abstract}

\section{INTRODUCTION}
Utilizing a powered hip exoskeleton to assist locomotion can reduce metabolic cost, augment human performance and restore abnormal gait \cite{siviy2023opportunities} \cite{improve}. While most researchers explored the benefits of lower limb exoskeleton by walking on the treadmill with periodic and constant assistance, this simple control strategy is not suitable for real applications \cite{Medrano2023TRO}. During daily walking, the human hip kinematics and kinetics varies according to the terrains \cite{Camargo2021data}. Therefore, how to provide appropriate assistance on different terrains is crucial for hip exoskeleton daily usage.

There are two main strategies for terrain adaptive assistance. One strategy is recognizing the type of terrain and then providing assistance based on the pre-defined trajectory/torque profile. This method is also called locomotion mode recognition \cite{Aaron2022LMR}. Common types of terrain include level walking (LW), stair ascent (SA), stair descent (SD), ramp ascent (RA) and ramp descent (RD) \cite{type}. Zhao et al. proposed a learning free method by terrain reconstruction and visual-inertial odometry \cite{learningfree}. Qian et al. designed a convolutional neural network (CNN) to classify the depth image captured by an RGB-D camera on the chest, which represented the upcoming terrain and achieved 98\% accuracy \cite{Qian2022RAL}. However, as there are many factors that could influence the gait pattern, such as ramp incline, stair height and walking speed, it's complicated to design the suitable profile considering all factors \cite{Sharma2023NSRE}.

The other strategy is directly forecasting the joint angle and then tracking the predicted results. Wei et al. utilized inertial measurement units (IMUs) to classify the locomotion mode and treated it as the prior knowledge to predict future joint angles, achieving normalized RMSE below 8.41\% for knee angle \cite{Wei2023Sensors}. Zhang et al. proposed a two staged forecasting framework, which first predicted the readings of IMUs then regressed the pose \cite{MoFCNet}. Due to the mechanical delay, only using IMUs shows sub-optimal performance during the locomotion transition. Bio-electrical signals like surface electromyography (sEMG) have been widely researched for human joint angle prediction, achieving great accuracy \cite{Sitole2023EMG} \cite{EMG2}. However, sensor's location, noise, inter-subject difference and the need for exposed skin makes it challenging for practical usage \cite{EMG3}. Fusing the terrain image with lower limb kinematics is a biomimetic strategy as human. Sharma et al. found that lower limb joint prediction accuracy could be improved by egocentric optical flow \cite{Sharma2022Optical}. In order to satisfy the practical usage, Tespa et al. designed a more lightweight neural network which can run on embedding device in real time, and the overall root mean square error (RMSE) was \(6.85^\circ\) for ankle angle and \(9.67^\circ\) for knee angle \cite{Tsepa2023ICRA}. 

The convenience of cameras and cross-subject invariant of terrain image make the visual information suitable for joint angle forecasting in daily use. However, it is still challenging to achieve the accurate prediction with the low computational cost of inference. To address this issue, we propound a solution with combination of more efficient data structure and advanced algorithm. For the data structure side, diving the whole time series into single strides may be more efficient than the widely used sliding window approach. It could simplify the problem into forecasting the joint's angle during one complete gait cycle, while sliding window approach neglects the periodicity of human's locomotion thus results in complicated pattern of leg's kinematics and corresponding terrain image. This simplification makes sense because most control strategies of exoskeletons are based on gait phase to solve the trade-off between the spatial guidance and the temporal freedom \cite{control}, and many researches show that the gait phase could be estimated accurately from IMUs signal \cite{Na2023GP} \cite{GP2} \cite{GP3}. 

For the algorithm side, simplified transformer mechanism according to the specific data structure may achieve better performance with lower computational cost. The transformer based architecture of neural network has shown better performance in many multi-modal tasks \cite{Xu2023transformer}, while it's still not widely used in joint angle forecasting field due to its higher computational cost than CNN and LSTM. Therefore, how to enhance the transformer architecture to promote it more efficient and accurate for joint angle prediction is another potential measure to make the integrated image information practical for exoskeletons in daily use.  

In order to prove our hypothesis, we proposed a stride-level sandwich data structure and design a multi-modal transformer network with sandwich fusion mechanism based on ViT for depth image \cite{RawViT} and PatchTST for IMUs time series \cite{PatchTST}, called SFTIK (Sandwich Fusion Transformer for Image and Kinematics). We shared a multi-person dataset which simultaneously collect egocentric RGB-D image and IMU signal with stride level notations. Finally we conducted comprehensive ablation experiments which suggest the proposed method achieving best performance in hip joint trajectory forecasting to date.

\section{METHODS}

\subsection{Data Collection}
\subsubsection{Subject}
Ten healthy and young subjects (6 males and 4 females, age 21.8 ± 1.9 years, height 172.4 ± 10.7 cm) were recruited to this experiment. Prior to their participation, all subjects were thoroughly briefed about the experimental procedures and the potential risks involved, ensuring informed consent was obtained.

\subsubsection{Hardware}
The images of terrain were captured by an RGB-D camera (Realsense D435, Intel), positioned at the chest, with a sampling rate set as 15 Hz to accommodate real-time color-depth alignment computations. Additionally, three nine-axis IMUs (CMP10A, Yahboom) were located at the posterior pelvis and the anterior surfaces of both thighs, each operating at a sampling rate of 100 Hz (Fig.\ref{fig_data_collect}.a). These sensors are interconnected via wires to a laptop, which is carried in a backpack. This laptop executes a Python-based data acquisition program, ensuring efficient data collection and processing.

\subsubsection{Experimental Protocol}
\begin{figure}[h]
	\centering
	\includegraphics[width=\linewidth]{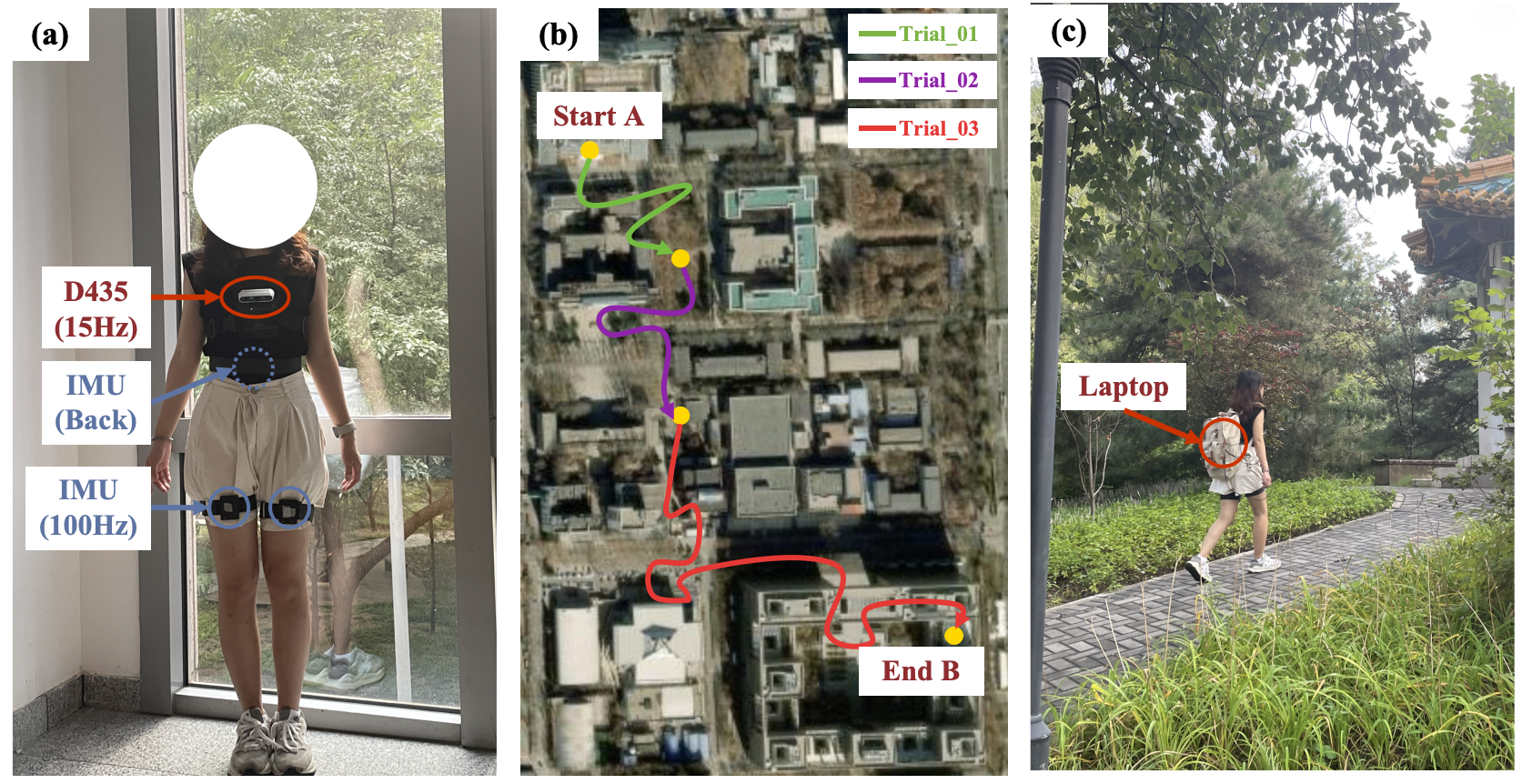}
	\caption{The experimental procedure for data collection. (a) shows the sensors used. A D435 camera is fixed on the chest. One IMU is positioned on the posterior pelvis and two IMUs are positioned on the left and right tights respectively. (b) illustrates the route employed for the data collection, encompassing three distinct trails. (c) presents an in-situ photo captured during the data collection. All the sensors mentioned before are linked to a laptop within a backpack. }
	\label{fig_data_collect}
\end{figure}

The experiment used a uniform walking route for all participants, starting from the third floor of building A and concluding on the second floor of build B, covering five common terrains (Fig.\ref{fig_data_collect}.b). The entire path was divided into three segments, each with a duration of roughly 10 minutes. At the end of each segment, participants were given a 3-minute rest period. After each rest, the IMUs were adjusted and calibrated by asking the participant stand still for about 10 seconds to set the IMU's sagittal plane angle as the bias \cite{Medrano2023TRO}. 
\subsection{Data Processing and Labeling}

 \begin{figure*}[!t]
	\centering
	\includegraphics[width=\linewidth]{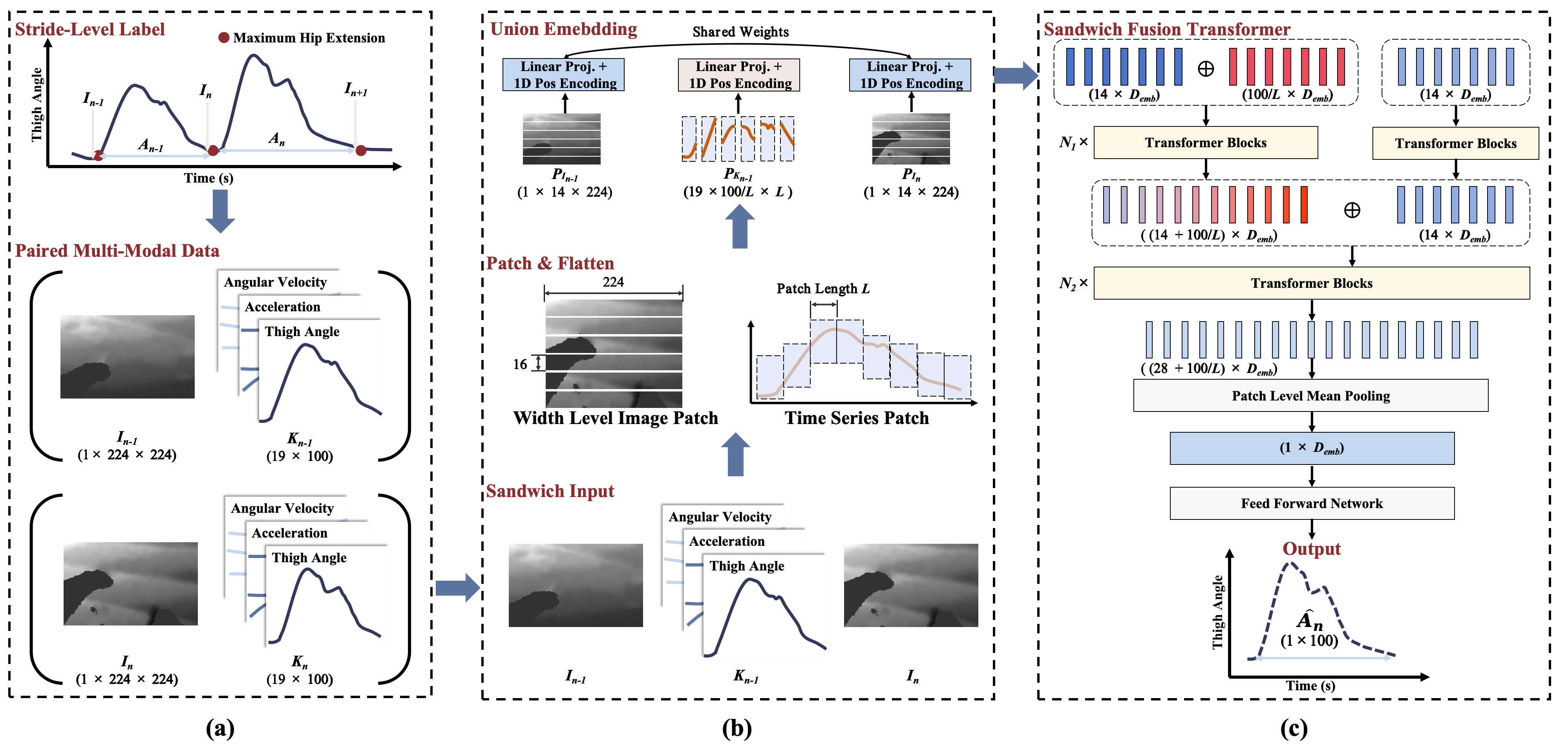}
	\caption{The framework of proposed SFTIK. (a) for each step, paired kinematic data and terrain image (\(K_n\), \(I_n\)) were constructed based on the maximum hip extension. (b) the inputs of SFTIK (\(K_{n-1}\), \(I_{n-1}\), \(I_n\)) were patched and embedded into the same dimension \(D_{emb}\). (c) the sandwich fusion mechanism first applied \(N_1\) layers of transformer blocks to learn the relationship between \(K_{n-1}\) and \(I_{n-1}\). Then \(N_2\) layers of transformer blocks were used to forecast the latent feature of thigh angle \(A_n\) according to \(I_n\) and the previously learned relationship. Finally, a combination of mean pooling and feed forward network produced the value of \(A_n\) .}
	\label{network}
\end{figure*}

During the data pre-processing, all channels of the IMU data were subjected to a second-order low-pass Butterworth filter with a cut-off frequency of 30 Hz to remove high frequency noise \cite{low_pass_filter}. For the RGB images, we applied standard normalization using the mean and standard deviation values from ImageNet. Depth images were processed by clipping the range to 0-5 meters and subsequently normalized by dividing by the maximum value (5m). Both RGB and depth images were then resized to 224x224. 

During the data labeling, each side of leg was treated separately. We utilized the on-board calculated IMU's sagittal plane angle as thigh angle and pelvis angle. The maximum hip extension (MHE) was computed as the defining characteristic gait event to segment the IMU data stream into individual strides \cite{Qian2022RAL}. The amount of valid strides was shown in Table \ref{table_amount}. For each stride, the IMU data were resampled to a uniform length of 100 points via interpolation. After that, the proximate terrain image preceding the MHE event was identified based on timestamp alignment and designated as the key-frame. The locomotion mode for each stride was then determined based on its corresponding key-frame.

\begin{table}[h]
\caption{Amount of valid strides on different terrains}
\label{table_amount}
\centering
\begin{tabular}{cccccc}
\toprule
LW & RA & RD & SA & SD & Whole \\
\midrule
19,865 & 2,360 & 2,189 & 1,987 & 2,001 & 28,402 \\
\bottomrule
\end{tabular}
\end{table}

\subsection{Architecture of proposed SFTIK}
\subsubsection{Problem Formulation}
We constructed a dataset comprising paired kinematic data and terrain image (\(K_n\), \(I_n\)) for each step. Specifically, the kinematics data 
 \(K_n\) with shape (19, 100), represent the thigh angle, tri-axial acceleration, and tri-axial angular velocity, recorded from the pelvis and both thighs' IMUs. The terrain image \(I_n\) with dimensions (1, 224, 224), represent the depth terrain image captured at the beginning of each gait cycle. Our objective is to leverage the model to predict thigh angle \(A_n\) of ensuing step given the kinematics data \(K_{n-1}\) from preceding step and terrain images (\(I_{n-1}\), \(I_n\)) from the current and preceding steps, as shown in (\ref{eq:Model_Formulation}).
 
\begin{equation}
\hat{A_n} = \mathcal{F}(K_{n-1}, I_{n-1}, I_n; \theta)
\label{eq:Model_Formulation}
\end{equation}
where $\hat{A_n}$ denotes the forecasting thigh angle, $\mathcal{F}$ refers to the model and $\theta$ refers to its parameters.

\subsubsection{Patchify and Union Embedding Block}
Typically, the image is patched using a 16 $\times$ 16 square kernel, which incurs significant computational costs, rendering it unsuitable for embedding devices used in exoskeleton. We observed that egocentric terrain images exhibit inherent structure, with the height direction consistently aligning with the human's facing direction. This observation suggests that terrain information mainly varies along the image's height direction. Therefore, we proposed a 224 × 16 width-level kernel for image patching to reduce the computational burden. The IMU time series data were segmented using a patch length of $L$ and a stride length of $L_s$. Subsequently, each patch was mapped to the uniform dimension $D_{emb}$ via linear projection, and was then enriched with 1-D learnable positional encoding. The weights of linear projection and positional encoding for both image $I_{n-1}$ and $I_n$ were shared.

\subsubsection{Sandwich Fusion Transformer}
The first fusion stage concatenated patches from the previous step's kinematic data $P_{K_{n-1}}$ and terrain image $P_{I_{n-1}}$ using $N_1$ layers of basic transformer block. In order to ensure symmetry, patches of current step's image $P_{I_{n}}$ were also processed through $N_1$ layers of the transformer block, as shown in (\ref{eq:Fusion 1.1}) and (\ref{eq:Fusion 1.2}).
\begin{equation}
{O_{IK_{n-1}}} = \mathcal{T}^{N_1}(P_{K_{n-1}}, P_{I_{n-1}};\theta_{IK})
\label{eq:Fusion 1.1}
\end{equation}
\begin{equation}
{P_{I'_{n}}} = \mathcal{T}^{N_1}( P_{I_{n-1}};\theta_{I_{N_1}})
\label{eq:Fusion 1.2}
\end{equation}
where $O_{IK_{n-1}}$ refers to the latent feature of the terrain image and kinematics from the previous stride, $P_{I'_{n}}$ refers to the latent feature of the terrain image from the current stride, $\mathcal{T}^{N_1}$ denotes $N_1$ layers of transformer block, and $\theta_{IK}$ and $\theta_{I_{N_1}}$ denote the parameters. Then the second fusion stage integrated $O_{IK_{n-1}}$ and $P_{I'_{n}}$ using $N_2$ layers of basic transformer block. We set the $N_1$ and $N_2$ as 6 for default. The amount of heads for all transformer block was set to be 12.

\begin{equation}
{L_{An}} = \mathcal{T}^{N_2}(O_{IK_{n-1}},{P_{I'_{n}};\theta_{N_2}})
\label{eq:Fusion 1.2}
\end{equation}

\begin{figure}[h]
	\centering
	\includegraphics[width=\linewidth]{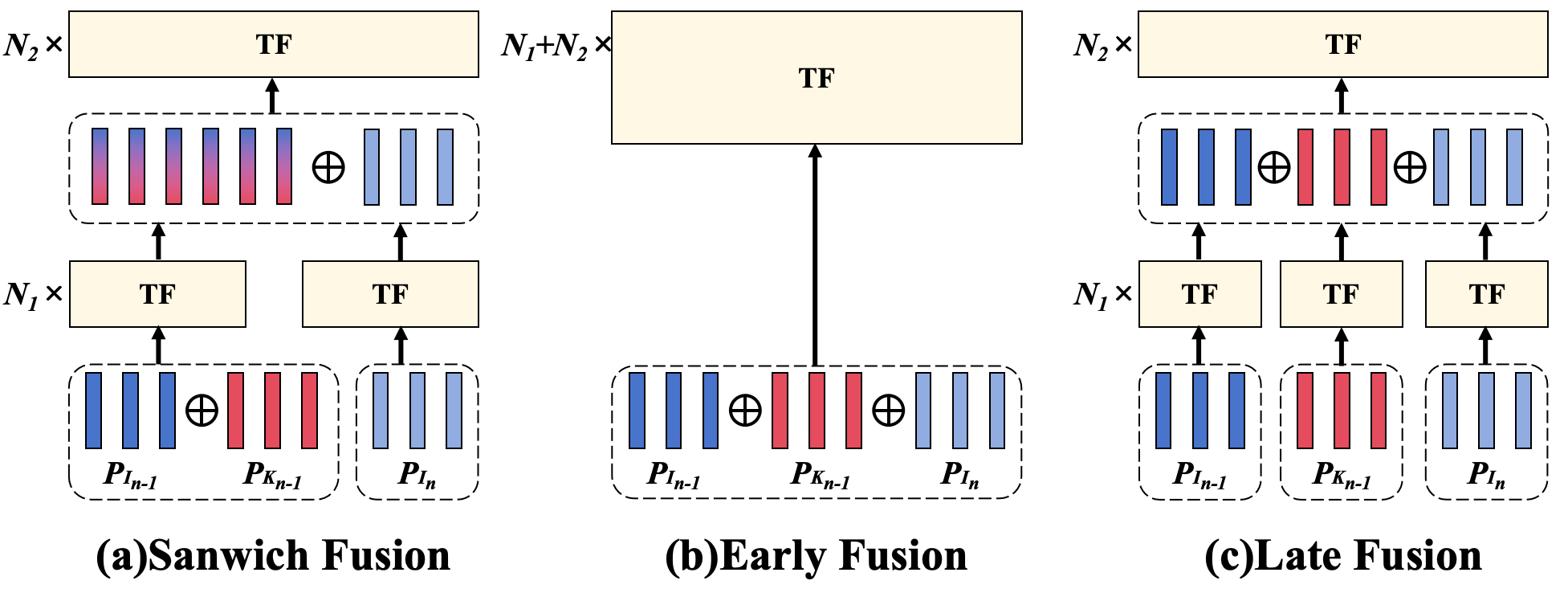}
	\caption{Comparative Illustration of Fusion Methods with Identical Depth. (a) shows the Sandwich Fusion method, showcasing the two-step concatenation process involving previous and current stride's terrain images and kinematic data. (b) illustrates the Early Fusion method, where all patches are concatenated prior to processing through transformer blocks. (c) shows the Late Fusion approach, highlighting the sequential application of transformer blocks to individual patches followed by a collective concatenation and subsequent transformer block processing.}
	\label{fig_fusion_method}
\end{figure}

This mechanism could be considered as forecasting the latent feature $L_{An}$ of thigh angle via current step's terrain image and the previous learned relationship $O_{IK_{n-1}}$ between terrain image and corresponding hip kinematics. The difference between the proposed sandwich fusion and commonly early/late fusion was shown in Fig.\ref{fig_fusion_method}.

After that, a combination of patch level mean pooling layer and two layers feed forward network (FFN) were used to project the latent feature of thigh angle into thigh angle time series with shape $(1 \times 100)$.
\subsubsection{Loss Function}
We adopted the mean square error (MSE) loss for time series forecasting task as the loss function, as shown in (\ref{eq:loss}).
\begin{equation}
L = \frac{1}{n} \sum_{i=1}^{n} ({\hat{A_n}_i} - {A_n}_i)^2
\label{eq:loss}
\end{equation}
where $n$ denotes the batch size, $\hat{A_n}$ denotes the forecasting thigh angle and $A_n$ denotes the ground truth.

\subsection{Training details}
The models were trained in a Linux computer with an Nvidia RTX 4090 GPU. The python version is 3.11.4, the pytorch version is 2.1.0, and the cuda verision is 11.8. During training, we set the batch size as 32, the epoch as 200, and the optimizer to be Adam. The learning rate is $2 \times 10^{-4}$ incorporating a warm-up strategy over 50 steps, starting with a ratio of 0.2 and employing cosine growth, subsequently transitioning to cosine annealing towards the end of the training.
\section{RESULTS}
The performance of every method was reported as root mean square error (RMSE) and Pearson's correlation coefficient (PCC) with standard deviation, and was tested with a 5-fold leave-one-out cross-validation (LOO-CV). In each fold of LOO-CV, data from seven subjects were designated for training, one for validation and two for testing.
\subsection{IMU Time Series Embedding}
\subsubsection{Comparison between sliding window and stride-level}
\begin{figure}[h]
	\centering
	\includegraphics[width=\linewidth]{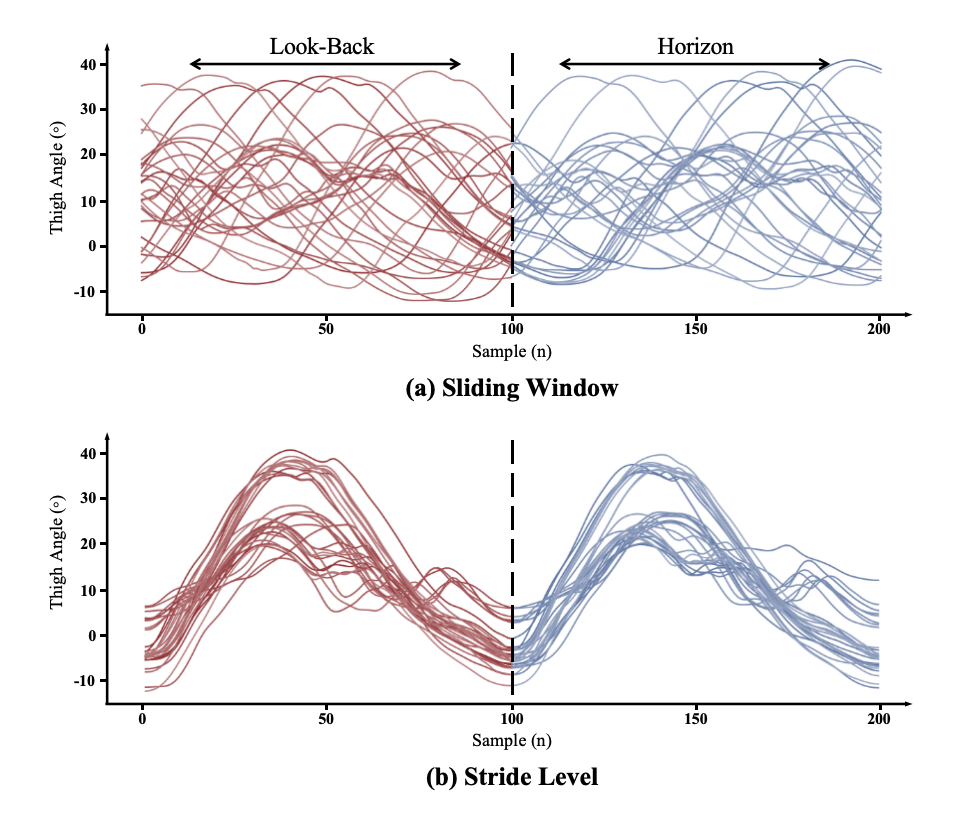}
	\caption{Time series pattern comparison between sliding window and stride level. (a) Sliding window method with look-back, prediction and stride length = 100. (b) Stride-level method with one gait cycle length = 100.}
	\label{fig_sl_vs_sw}
\end{figure}

We visualized 30 input-output paired series of thigh angles used for network training in Fig.\ref{fig_sl_vs_sw}. The sliding window employed for both the look-back and prediction encompasses 100 samples. The series length of stride level approach was also standardized to 100 samples by interpolation.

To further substantiate the efficacy of the stride-level methodology, the PatchTST model was employed as the foundational framework to forecast the thigh angle with both dividing method. The results are shown in Table \ref{table_sl_vs_sw}.

\begin{table}[h]
\centering
\caption{Forecasting Results for Sliding Window and Stride-level}
\label{table_sl_vs_sw}
\begin{tabular}{ccc}
\toprule
Method & RMSE(\textdegree) & PCC\\
\midrule
Sliding Window & 5.034 \textpm\ 0.374 & 0.934 \textpm\ 0.008 \\
Stride Level & \textbf{3.369 \textpm\ 0.273} & \textbf{0.978 \textpm\ 0.002} \\
\bottomrule
\end{tabular}
\end{table}

\subsubsection{Patch Length}
We conducted experiments to find the influence of patch length $L$ and stride length $S$ on SFTIK. The results are shown in Table \ref{tabel_patch_len}.

\begin{table}[h]
\centering
\caption{Forecasting Results for Patch Length Experiment}
\label{tabel_patch_len}
\begin{tabular}{ccccccc}
\toprule
Metric & $L=5, S=5$ & $L=10, S=10$  & $L=20, S=20$ \\
\midrule
LW\textsubscript{RMSE}(\textdegree) & \textbf{3.015 \textpm\ 0.447} & 3.025 \textpm\ 0.435 & 3.293 \textpm\ 0.902 \\
RA\textsubscript{RMSE}(\textdegree) & 3.456 \textpm\ 0.678 & \textbf{3.376 \textpm\ 0.431} & 3.529 \textpm\ 0.683 \\
RD\textsubscript{RMSE}(\textdegree) & 3.025 \textpm\ 0.517 & \textbf{2.922 \textpm\ 0.408} & 3.184 \textpm\ 0.758 \\
SA\textsubscript{RMSE}(\textdegree) & 4.499 \textpm\ 0.746 & \textbf{4.281 \textpm\ 0.851} & 4.703 \textpm\ 0.795 \\
SD\textsubscript{RMSE}(\textdegree) & 3.691 \textpm\ 0.786 & 3.633 \textpm\ 0.969 & \textbf{3.293 \textpm\ 0.902} \\
Avg.\textsubscript{RMSE}(\textdegree) & 3.537 \textpm\ 0.831 & \textbf{3.445 \textpm\ 0.804} & 3.795 \textpm\ 1.092 \\
PCC & 0.969 \textpm\ 0.029 & \textbf{0.971 \textpm\ 0.025} & 0.967 \textpm\ 0.030 \\
\bottomrule
\end{tabular}
\end{table}

\subsection{Terrain Image Embedding}

The impact of image modalities and patchify strategies on the performance of our proposed terrain image embedding block was evaluated by ablation experiment. First we applied width-level patch to deal with RGB, RGB-D and Depth image. After that we tested height-level and square patch on depth image. The results are shown in Table \ref{tabel_img}. The computational complexity of each embedding method was quantified with 6 layers of transformer block and reported as flops.

\begin{figure}[h]
	\centering
	\includegraphics[width=\linewidth]{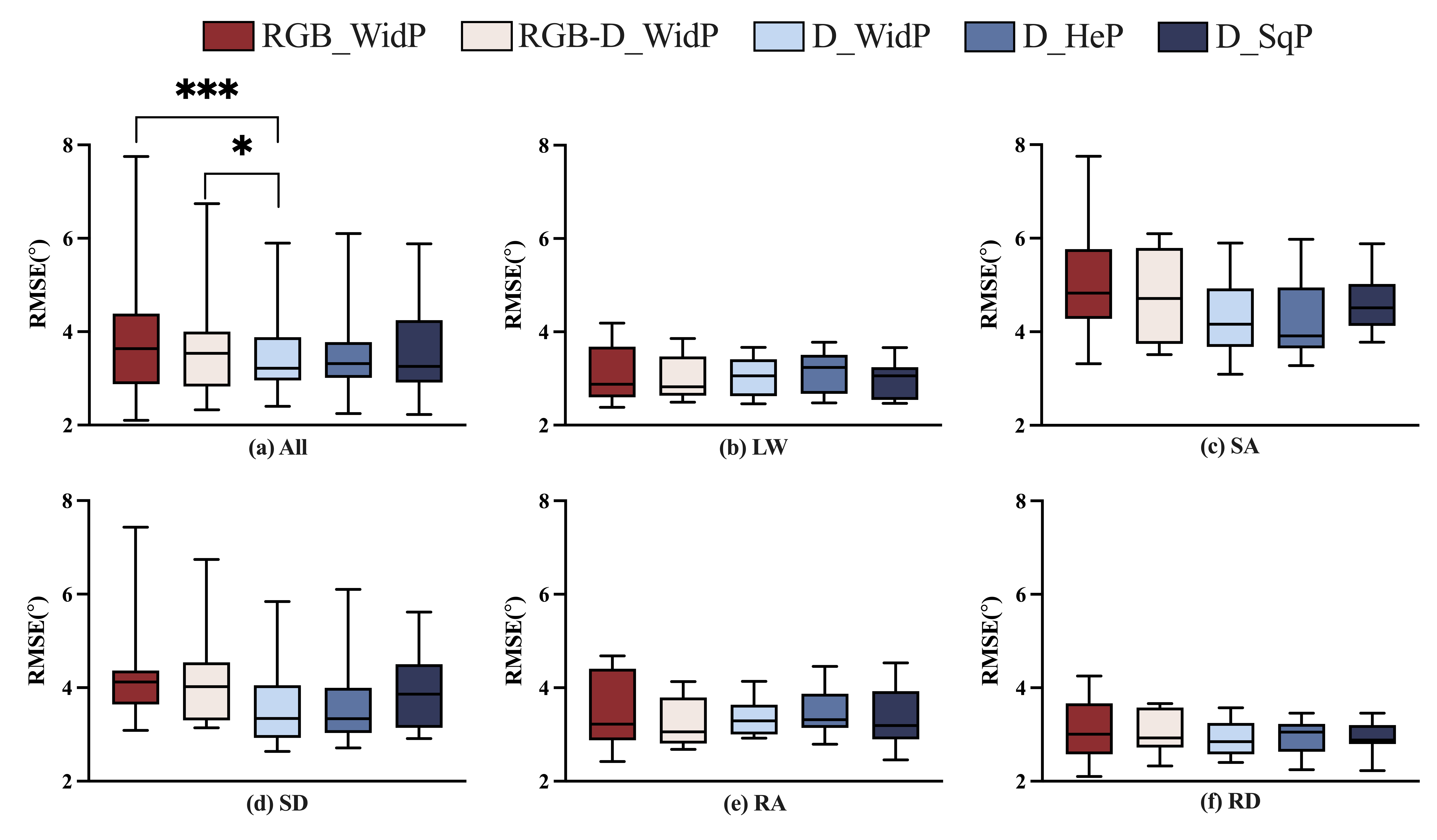}
    \captionsetup{justification=centering}
	\caption{Box-plot of RMSE across modality and patch methods of image  (* means p \textless \ 0.05, ** means p \textless \ 0.01, and *** means p \textless \ 0.001).}
	\label{fig_img_encoding}
\end{figure}

\begin{table}[h]
\centering
\caption{Forecasting Results for Image Modality and Patch Method Ablation}
\label{tabel_img}
\begin{tabular}{cccccc}
\toprule
Method & RMSE(\textdegree) & PCC & Flops(G) \\
\midrule
RGB\_WidP & 3.796 \textpm\ 1.204 & 0.965 \textpm\ 0.032 & 0.71 \\
RGB-D\_WidP & 3.634 \textpm\ 1.003 & 0.968 \textpm\ 0.029 & 0.75 \\
D\_WidP & \textbf{3.445 \textpm\ 0.804} & \textbf{0.971 \textpm\ 0.025} & \textbf{0.63} \\
D\_HeP & 3.502 \textpm\ 0.815 & 0.971 \textpm\ 0.023 & \textbf{0.63} \\
D\_SqP & 3.558 \textpm\ 0.879 & 0.969 \textpm\ 0.029 & 8.37 \\
\bottomrule
\end{tabular}
\end{table}
To further compare the performance across different locomotion, we visualized the forecasting RMSE of thigh angle by subject (n=10), as shown in Fig.\ref{fig_img_encoding}. Meanwhile, we conducted Wilcoxon rank-sum paired test to statistically examine the overall significance.

\subsection{Vision and Kinematics Fusion}

We compared proposed sandwich fusion mechanism with early fusion and late fusion. To ensure a fair comparison, the layers of transformer blocks were equal across fusion methods. We also conducted information ablation study without terrain image $I_{n-1}$ of preceding step and without the kinematics data $K_{n-1}$. The results are shown in Table \ref{tabel_fusion}.

\begin{figure}[h]
	\centering
	\includegraphics[width=\linewidth]{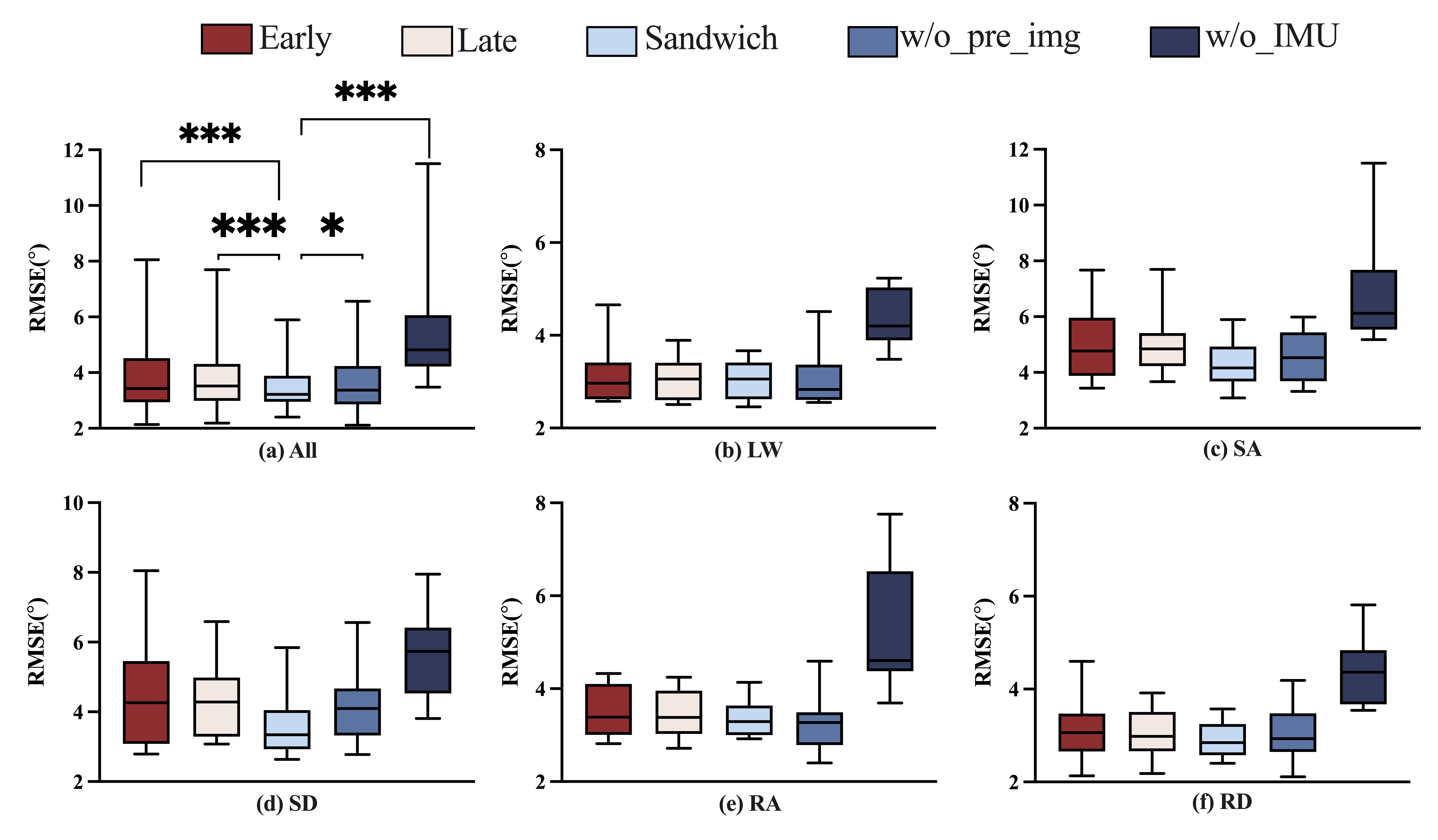}
    \captionsetup{justification=centering}
	\caption{Box-plot of RMSE across fusion methods and information ablation  (* means p \textless \ 0.05, ** means p \textless \ 0.01, and *** means p \textless \ 0.001).}
	\label{fig_fusion_result}
\end{figure}

\begin{table}[h]
\centering
\caption{Forecasting Results for Fusion Method and Information Ablation}
\label{tabel_fusion}
\begin{tabular}{ccccc}
\toprule
Method & RMSE(\textdegree) & PCC \\
\midrule
Early & 3.852 \textpm \ 1.274 & 0.966 \textpm \ 0.031  \\
Late & 3.778 \textpm \ 1.098 & 0.965 \textpm \ 0.031  \\
Sandwich & \textbf{3.445 \textpm \ 0.804} & \textbf{0.971 \textpm \ 0.025}  \\
w/o\_pre\_img & 3.617 \textpm \ 1.011 & 0.968 \textpm \ 0.031  \\
w/o\_IMU & 5.263 \textpm \ 1.514 & 0.954 \textpm \ 0.040  \\
\bottomrule
\end{tabular}
\end{table}
We also visualized the RMSE across locomotion and conducted statistically examining, as shown in Fig.\ref{fig_fusion_result}.
\subsection{Hyper-parameters Tuning}
The hyper-parameters of the SFTIK are the embedding dimension \(D_{emb}\), depth of the first fusion block \(N_1\) and depth of the second fusion block \(N_2\). The tuning results are shown in Table \ref{tabel_hyper}.

\begin{table}[h]
\centering
\caption{Forecasting Results for Hyper-Parameters Tuning}
\label{tabel_hyper}
\begin{tabular}{ccccc}
\toprule
\(D_{emb}\) & \(N_1\) & \(N_2\) & RMSE(\textdegree) & PCC \\
\midrule
768 & 6 & 6 & \textbf{3.445 \textpm\ 0.804} & \textbf{0.971 \textpm\ 0.025} \\
576 & 6 & 6 & 3.478 \textpm\ 0.784 & 0.971 \textpm\ 0.025 \\
384 & 6 & 6 & 3.605 \textpm\ 0.882 & 0.970 \textpm\ 0.026 \\
192 & 6 & 6 & 3.583 \textpm\ 0.818 & 0.969 \textpm\ 0.025 \\
768 & 3 & 9 & 3.566 \textpm\ 0.893 & 0.970 \textpm\ 0.026 \\
768 & 9 & 3 & 3.537 \textpm\ 0.852 & 0.970 \textpm\ 0.027 \\
768 & 3 & 3 & 3.561 \textpm\ 0.857 & 0.970 \textpm\ 0.026 \\
\bottomrule
\end{tabular}
\end{table}

\subsection{Baseline comparison}

We compared our proposed SFTIK with original ViT-PatchTST network, ResNet-LSTM and MobileNet-MLP \cite{Tsepa2023ICRA}. The results are shown in Table \ref{tabel_baseline}.

\begin{table}[h]
\centering
\caption{Forecasting Results for Baseline Comparison}
\label{tabel_baseline}
\begin{tabular}{cccccc}
\toprule
Model & RMSE(\textdegree) & PCC & Flops(G) \\
\midrule
SFTIK(ours) & \textbf{3.445 \textpm\ 0.804} & \textbf{0.971 \textpm\ 0.025} & 3.31 \\
ViT-PatchTST & 3.724 \textpm\ 0.881 & 0.968 \textpm\ 0.031 & 33.7 \\
MobileNet-MLP & 3.481 \textpm\ 0.827 & 0.970 \textpm\ 0.029 & \textbf{1.12} \\
ResNet-LSTM & 3.515 \textpm \ 0.847 & 0.969 \textpm\ 0.029 & 28.9 \\
\bottomrule
\end{tabular}
\end{table}

\section{DISCUSSION}

The proposed paired image and IMU time series data based on stride provides a lucid and simplistic way to describe the relationship between locomotion and terrain. First, it could reduce the possible patterns of input kinematics, making the model easier to learn and converge. The sliding window method captures intricate patterns based on the specific phase range, requiring complex network and massive data to learn the patterns. In contrast, stride-level analysis encapsulated solely the leg's motion from the beginning to the end of a single gait cycle. The results show applying stride-level analysis could reduce the RMSE by 33.08\%. Second, the paired image only represents the terrain at the beginning of the gait cycle, making the model more accurate and faster to learn and converge. The image captured on the random gait phase may confuse the model with human body's interference and terrain transition.

The large computational demands of the transformer model pose challenges for its real-time application on edge devices, such as exoskeletons. We noticed that image processing requires more computational resources than time series analysis, then proposed width level patch to reduce cost. Contrary to typical images that contain information in all directions, the terrain images we collected are egocentric, with most of the relevant information concentrated in the width direction. Implementing either width or height level patch could reduce the flops around 10 times compared to the standard square patch and achieve better performance.

Thoroughly reviewing of the preceding stride, while maintaining the chronological order, could enhance the forecasting performance. The proposed sandwich fusion mechanism could first learn the latent relationship between the terrain image and kinematics, then infer the ensuing thigh angle using the learned relationship and ensuing terrain image. Ablation of the chronological fusion strategy such as the early or late fusion would deteriorate the forecasting RMSE around 8.73-10.49\%. Each part of the proposed sandwich data is important, omitting the terrain image or the IMU time series of the preceding stride would deteriorate the forecasting RMSE by 4.67\% and 34.49\%, separately.

One limitation of our study is that we only forecast the profile between thigh angle and gait phase of next stride. It still needs a gait phase estimation model for real applications. So our method differs from typical trajectory forecasting approaches, as it generates a terrain adaptive trajectory profile for a hip exoskeleton through forecasting. Besides, we evaluate the performance according to subject walking the same route. The generalization ability on new terrain needs further research.

\section{CONCLUSION}
In order to plan the trajectory for controlling a hip exoskeleton's on different terrains, we proposed the SFTIK to predict the thigh angle of the upcoming stride based on terrain images and IMU time series data. A comprehensive dataset with 28,402 valid strides featuring real-world walking scenarios, inclusive of hip kinematics and terrain images, was collected and annotated for performance evaluation. Ablation studies demonstrated that the sandwich structure of the input data and its corresponding fusion mechanism significantly enhance predictive performance. The width-level patchify approach, tailored for egocentric terrain images, could reduce computational complexity more than 10 times while improve performance than ordinary square patchify by 3.18\%. Overall, the proposed SFTIK outperforms all baseline methods, achieving a computational efficiency of 3.31 G Flops, and RMSE of 3.445 \textpm \ 0.804\textdegree \ and PCC of 0.971 \textpm\ 0.025.

\bibliographystyle{IEEEtran}
\bibliography{references}

\begin{thebibliography}{10}
\providecommand{\url}[1]{#1}
\csname url@rmstyle\endcsname
\providecommand{\newblock}{\relax}
\providecommand{\bibinfo}[2]{#2}
\providecommand\BIBentrySTDinterwordspacing{\spaceskip=0pt\relax}
\providecommand\BIBentryALTinterwordstretchfactor{4}
\providecommand\BIBentryALTinterwordspacing{\spaceskip=\fontdimen2\font plus
\BIBentryALTinterwordstretchfactor\fontdimen3\font minus \fontdimen4\font\relax}
\providecommand\BIBforeignlanguage[2]{{%
\expandafter\ifx\csname l@#1\endcsname\relax
\typeout{** WARNING: IEEEtran.bst: No hyphenation pattern has been}%
\typeout{** loaded for the language `#1'. Using the pattern for}%
\typeout{** the default language instead.}%
\else
\language=\csname l@#1\endcsname
\fi
#2}}

\bibitem{siviy2023opportunities}
C.~Siviy, L.~M. Baker, B.~T. Quinlivan, F.~Porciuncula, K.~Swaminathan, L.~N. Awad, and C.~J. Walsh, ``Opportunities and challenges in the development of exoskeletons for locomotor assistance,'' \emph{Nat. Biomed. Eng.}, vol.~7, no.~4, pp. 456--472, 2023.

\bibitem{improve}
C.~Jayaraman, K.~R. Embry, C.~K. Mummidisetty, Y.~Moon, M.~Giffhorn, S.~Prokup, B.~Lim, J.~Lee, Y.~Lee, M.~Lee, \emph{et~al.}, ``Modular hip exoskeleton improves walking function and reduces sedentary time in community-dwelling older adults,'' \emph{J. Neuroeng. Rehabil.}, vol.~19, no.~1, p. 144, 2022.

\bibitem{Medrano2023TRO}
R.~L. Medrano, G.~C. Thomas, C.~G. Keais, E.~J. Rouse, and R.~D. Gregg, ``Real-time gait phase and task estimation for controlling a powered ankle exoskeleton on extremely uneven terrain,'' \emph{IEEE Trans. Robot.}, vol.~39, no.~3, pp. 2170--2182, 2023.

\bibitem{Camargo2021data}
J.~Camargo, A.~Ramanathan, W.~Flanagan, and A.~Young, ``A comprehensive, open-source dataset of lower limb biomechanics in multiple conditions of stairs, ramps, and level-ground ambulation and transitions,'' \emph{J. Biomech.}, vol. 119, p. 110320, 2021.

\bibitem{Aaron2022LMR}
I.~Kang, D.~D. Molinaro, G.~Choi, J.~Camargo, and A.~J. Young, ``Subject-independent continuous locomotion mode classification for robotic hip exoskeleton applications,'' \emph{IEEE Trans. Biomed. Eng.}, vol.~69, no.~10, pp. 3234--3242, 2022.

\bibitem{type}
M.~Sharifi-Renani, M.~H. Mahoor, and C.~W. Clary, ``Biomat: An open-source biomechanics multi-activity transformer for joint kinematic predictions using wearable sensors,'' \emph{Sensors}, vol.~23, no.~13, p. 5778, 2023.

\bibitem{learningfree}
S.~Zhao, Z.~Yu, Z.~Wang, H.~Liu, Z.~Zhou, L.~Ruan, and Q.~Wang, ``A learning-free method for locomotion mode prediction by terrain reconstruction and visual-inertial odometry,'' \emph{IEEE Trans. Neural Syst. Rehabil. Eng.}, vol.~31, pp. 3895--3905, 2023.

\bibitem{Qian2022RAL}
Y.~Qian, Y.~Wang, C.~Chen, J.~Xiong, Y.~Leng, H.~Yu, and C.~Fu, ``Predictive locomotion mode recognition and accurate gait phase estimation for hip exoskeleton on various terrains,'' \emph{IEEE Robot. Autom. Lett.}, vol.~7, no.~3, pp. 6439--6446, 2022.

\bibitem{Sharma2023NSRE}
A.~Sharma and E.~Rombokas, ``Optimizing representations of multiple simultaneous attributes for gait generation using deep learning,'' \emph{IEEE Trans. Neural Syst. Rehabil. Eng.}, vol.~31, pp. 2296--2305, 2023.

\bibitem{Wei2023Sensors}
B.~C. Wei, C.~Z. Yi, S.~P. Zhang, H.~Guo, J.~F. Zhu, Z.~Ding, and F.~Jiang, ``Taking locomotion mode as prior: One algorithm-enabled gait events and kinematics prediction on various terrains,'' \emph{IEEE Sens. J.}, vol.~23, no.~12, pp. 13\,072--13\,083, 2023.

\bibitem{MoFCNet}
X.~Zhang, H.~Zhang, J.~Hu, J.~Deng, and Y.~Wang, ``Motion forecasting network (mofcnet): Imu-based human motion forecasting for hip assistive exoskeleton,'' \emph{IEEE Robot. Autom. Lett.}, vol.~8, no.~9, pp. 5783--5790, 2023.

\bibitem{Sitole2023EMG}
S.~P. Sitole and F.~C. Sup, ``Continuous prediction of human joint mechanics using emg signals: A review of model-based and model-free approaches,'' \emph{IEEE Trans. Med. Robot. Bionics}, vol.~5, no.~3, pp. 528--546, 2023.

\bibitem{EMG2}
Y.~Wang, X.~Cheng, L.~Jabban, X.~Sui, and D.~Zhang, ``Motion intention prediction and joint trajectories generation toward lower limb prostheses using emg and imu signals,'' \emph{IEEE Sens. J.}, vol.~22, no.~11, pp. 10\,719--10\,729, 2022.

\bibitem{EMG3}
H.~Xu and A.~Xiong, ``Advances and disturbances in semg-based intentions and movements recognition: A review,'' \emph{IEEE Sens. J.}, vol.~21, no.~12, pp. 13\,019--13\,028, 2021.

\bibitem{Sharma2022Optical}
A.~Sharma and E.~Rombokas, ``Improving imu-based prediction of lower limb kinematics in natural environments using egocentric optical flow,'' \emph{IEEE Trans. Neural Syst. Rehabil. Eng.}, vol.~30, pp. 699--708, 2022.

\bibitem{Tsepa2023ICRA}
O.~Tsepa, R.~Burakov, B.~Laschowski, A.~Mihailidis, and Ieee, ``Continuous prediction of leg kinematics during walking using inertial sensors, smart glasses, and embedded computing,'' in \emph{Proc. IEEE Int. Conf. Robot. Autom.}, 2023, pp. 10\,478--10\,482.

\bibitem{control}
M.~Shushtari, H.~Dinovitzer, J.~Weng, and A.~Arami, ``Ultra-robust real-time estimation of gait phase,'' \emph{IEEE Trans. Neural Syst. Rehabil. Eng.}, vol.~30, pp. 2793--2801, 2022.

\bibitem{Na2023GP}
J.~Na, H.~Kim, G.~Lee, and W.~Nam, ``Deep domain adaptation, pseudo-labeling, and shallow network for accurate and fast gait prediction of unlabeled datasets,'' \emph{IEEE Trans. Neural Syst. Rehabil. Eng.}, vol.~31, pp. 2448--2456, 2023.

\bibitem{GP2}
Y.~Guo, Y.~Hutabarat, D.~Owaki, and M.~Hayashibe, ``Speed-variable gait phase estimation during ambulation via temporal convolutional network,'' \emph{IEEE Sens. J.}, vol.~24, no.~4, pp. 5224--5236, 2024.

\bibitem{GP3}
I.~Kang, D.~D. Molinaro, S.~Duggal, Y.~Chen, P.~Kunapuli, and A.~J. Young, ``Real-time gait phase estimation for robotic hip exoskeleton control during multimodal locomotion,'' \emph{IEEE Robot. Autom. Lett.}, vol.~6, no.~2, pp. 3491--3497, 2021.

\bibitem{Xu2023transformer}
P.~Xu, X.~T. Zhu, and D.~A. Clifton, ``Multimodal learning with transformers: A survey,'' \emph{IEEE Trans. Pattern Anal. Mach. Intell.}, vol.~45, no.~10, pp. 12\,113--12\,132, 2023.

\bibitem{RawViT}
A.~Dosovitskiy, L.~Beyer, A.~Kolesnikov, D.~Weissenborn, X.~Zhai, T.~Unterthiner, M.~Dehghani, M.~Minderer, G.~Heigold, and S.~Gelly, ``An image is worth 16x16 words: Transformers for image recognition at scale,'' in \emph{Proc. Int. Conf. Learn. Represent.}, 2020.

\bibitem{PatchTST}
Y.~Nie, N.~H. Nguyen, P.~Sinthong, and J.~Kalagnanam, ``A time series is worth 64 words: Long-term forecasting with transformers,'' in \emph{Proc. Int. Conf. Learn. Represent.}, 2023.

\bibitem{low_pass_filter}
F.~Weigand, A.~H{\"o}hl, J.~Zeiss, U.~Konigorski, and M.~Grimmer, ``Continuous locomotion mode recognition and gait phase estimation based on a shank-mounted imu with artificial neural networks,'' in \emph{IEEE/RSJ Proc. Int. Conf. Intell. Robots Syst.}, 2022, pp. 12\,744--12\,751.

\end{thebibliography}

\end{document}